\documentclass[10pt,twocolumn,letterpaper]{article}

\usepackage{iccv}              

\definecolor{iccvblue}{rgb}{0.21,0.49,0.74}
\usepackage[pagebackref,breaklinks,colorlinks,allcolors=iccvblue]{hyperref}

\def\paperID{6353} 
\def\confName{ICCV}
\def\confYear{2025}

\title{MultiNeRF: Multiple Watermark Embedding for Neural Radiance Fields}

\author{
Yash Kulthe$^{1}$ \quad Andrew Gilbert$^{1}$ \quad John Collomosse$^{1,2}$ \\
$^{1}$CVSSP, University of Surrey, UK \quad\quad
$^{2}$Adobe Research \\
{\tt\small \{y.kulthe, a.gilbert, j.collomosse\}@surrey.ac.uk} \quad
{\tt\small collomos@adobe.com}
}

\begin{document}

\twocolumn[{%
\renewcommand\twocolumn[1][]{#1}%
\maketitle
\vspace{-15pt} 
   \includegraphics[width=1\linewidth]{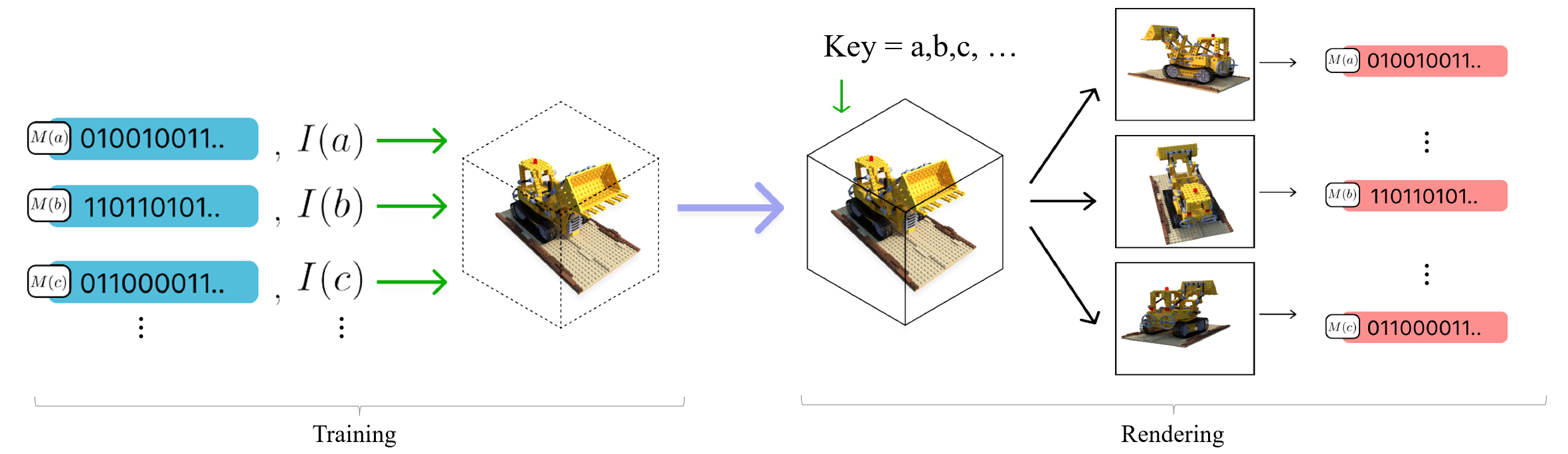}
    Figure 1. MultiNeRF embeds multiple watermarks within the representation learned by a NeRF model (TensoRF) at training time.  Watermarks are keyed by a unique ID specified at rendering time to trigger the embedding of the watermark into the image independent of the viewing position.
    \vspace{9.5pt}
    \label{fig:teaser}
}]

\setcounter{figure}{1}

\begin{abstract}
We present \textbf{MultiNeRF}\footnote{This research was funded by DECaDE under the UKRI Grant EP/T022485/1.}, a 3D watermarking method that embeds multiple uniquely keyed watermarks within images rendered by a single Neural Radiance Field (NeRF) model, whilst maintaining high visual quality. Our approach extends the TensoRF NeRF model by incorporating a dedicated watermark grid alongside the existing geometry and appearance grids. This extension ensures higher watermark capacity without entangling watermark signals with scene content. We propose a FiLM-based conditional modulation mechanism that dynamically activates watermarks based on input identifiers, allowing multiple independent watermarks to be embedded and extracted without requiring model retraining. MultiNeRF is validated on the NeRF-Synthetic and LLFF datasets, with statistically significant improvements in robust capacity without compromising rendering quality. By generalizing single-watermark NeRF methods into a flexible multi-watermarking framework, MultiNeRF provides a scalable solution for 3D content attribution.
\end{abstract}

\section{Introduction}
\label{sec:intro}

Neural Radiance Fields (NeRFs) \citep{mildenhall2021nerf} are a powerful high-fidelity 3D scene representation method, enabling photorealistic novel view synthesis. Their uses include online gaming, immersive experiences, and large-scale metaverse environments \citep{fabra2024application, zhang2024moving, li2022rt}. However, this also introduces new challenges in intellectual property (IP) protection, as NeRF models, whether representing products, scenes, or avatars, can be expensive to produce and be easily shared or leaked, Establishing the provenance of a NeRF model is crucial for asserting ownership and rights, helping to mitigate these risks, and even opening up novel compensation frameworks for their reuse \citep{collomosse2024authenticity}.

Digital watermarking has been a cornerstone of IP protection for visual media, including images \citep{zhu2018hidden,fernandez2022watermarking,bui2023trustmark} and video \citep{fernandez2024video}. However, these methods fall short in the NeRF context because they protect only the 2D outputs (the rendered images) rather than the underlying 3D representation itself. Recent work has focused on watermarking models directly, for example, text-to-image diffusion models \cite{fernandez2023stable} and NeRFs \citep{luo2023copyrnerf,zhang2024gs,jang2024waterf,song2024protecting}. Embedding watermark signals into the NeRF representation ensures persistent identification of models even under novel rendering views.

This paper introduces MultiNeRF, a framework for conditionally embedding {\em multiple} watermarks within a NeRF model. Existing NeRF watermarking techniques are limited to encoding a single watermark within the rendered image, typically with low capacity (\eg 16--48 bits). Even within the broader image watermarking literature \cite{zhu2018hidden,tancik2020stegastamp,fernandez2022watermarking,bui2023trustmark}, it is difficult to surpass capacities of this magnitude order (\ie $< 100$ bits) whilst maintaining acceptable visual quality for creative use cases.  However, this is insufficient even to accommodate a URL \eg to an end-user license agreement.
NeRF raises the intriguing possibility of embedding multiple distinct watermarks within a single model, each of which may bear only small capacities but which, taken together, can encode higher payload capacities.  Further, multiple independently keyed watermarks admit scenarios requiring multiple licenses or stakeholders (Fig. 1).  For example, in collaborative environments such as co-developed metaverse worlds, different contributors may need distinct watermarks to establish ownership or track usage.  Our technical contributions are:

\begin{enumerate}
\item {{\bf Watermark Grid}. We introduce a dedicated watermark grid alongside the existing geometry and appearance grids of the learned NeRF representation. This grid prevents the entanglement of the encoded watermark with scene content, improving the watermark capacity and preserving the rendering quality whilst adding only a small model size overhead. } 

\item {{\bf Conditional Modulation}. We introduce FiLM-based modulation \cite{perez2018film}, applying a lightweight input embedding network to encode watermark-specific identifiers and dynamically control the activation of watermark features. This enables the conditional rendering of multiple watermarks within the same NeRF model.}

\end{enumerate}

\cref{fig:arch2} illustrates our architecture to achieve multiple watermark embedding. We train the modified NeRF model end-to-end,  using a pre-trained HiDDeN decoder \citep{zhu2018hidden} as the base model for watermark retrieval. To ensure robustness, we augment training with differentiable noise sources. We later show (\cref{sec:eval}) our approach to deliver statistically significant improvements in capacity (\ie the ability to store many watermarks with high bit accuracy),  without substantial quality change.  We demonstrate this for two standard NeRF datasets of multiple scenes (NeRF-Synthetic \cite{mildenhall2021nerf} and LLFF \cite{mildenhall2019local}), thus extending NeRF watermarking into a more flexible paradigm suitable for real-world collaborative and commercial settings. 

\section{Related Work}
\label{sec:formatting}

\noindent \textbf{Visual Watermarking and Media Provenance.}
Traditional watermarking techniques embed information in the spatial \cite{taha2022high, ghazanfari2011lsb++} or frequency domains \cite{navas2008dwt, li2007steganographic, pevny2010using}. Similarly, recent works have explored steganography in images \cite{baluja2017hiding}, NeRFs \cite{li2023steganerf} and 3D Gaussian Splatting \cite{liinstantsplamp}. Deep watermarking methods such as HiDDeN \cite{zhu2018hidden}, StegaStamp \cite{tancik2020stegastamp}, RoSteALS \cite{bui2023rosteals}, SSL \cite{fernandez2022watermarking}, InvisMark \cite{xu2024invismark} and TrustMark \cite{bui2023trustmark} use learned embedding networks to improve imperceptibility and robustness against common transformations.  Watermarking has been used to help trace digital content's provenance (including ownership and rights) in combination with cryptographic metadata standards \eg C2PA. These standards attach signed metadata to digital assets, but metadata is frequently removed during redistribution. Watermarking provides a complementary approach by embedding identifiers directly into content, allowing provenance to persist even when metadata is stripped \cite{collomosse2024authenticity}; MultiNeRF explores NeRF watermarking to achieve the same goal.

\noindent \textbf{Watermarking NeRF models.} 
While deep watermarking has been extensively studied for 2D images and videos, its application to 3D generative models, such as NeRFs, is relatively new. Conventional 2D watermarking fails to ensure persistence across novel viewpoints, leading to recent research into NeRF watermarking methods \cite{luo2023copyrnerf, zhang2024gs, jang2024waterf, song2024protecting}. CopyRNeRF \cite{luo2023copyrnerf} embeds watermark signals in the color feature field to ensure extraction from arbitrary viewpoints. In contrast, NeRFProtector \cite{song2024protecting} focuses on embedding a watermark from the start of training a NeRF scene. WateRF \cite{jang2024waterf} introduces a frequency-based embedding, enhancing resilience to noise and compression. Existing NeRF watermarking methods are constrained by limited capacity and the inability to handle multiple watermarks, which we address with MultiNeRF.

\noindent \textbf{Watermarking in Generative Models.}  
With the rise of generative models in images \cite{ramesh2022hierarchical}, video \cite{ho2022imagen}, and 3D asset creation \cite{mildenhall2021nerf}, protecting model IP has become a key concern. Model provenance methods include the watermarking of training data \cite{asnani2024promark, sablayrolles2020radioactive}, model fingerprinting \cite{zhang2021deep}, and proactive tagging techniques \cite{wang2021faketagger}.  Recent work in diffusion model watermarking, such as Stable Signature \cite{fernandez2023stable}, introduces an in-model watermarking approach that fine-tunes the decoder of a latent diffusion model to embed robust identifiers directly into generated images. Other methods, such as Tree-Ring \cite{wen2023tree}, inject patterns into the noise initialization step of diffusion models.  MultiNeRF extends these approaches to 3D by embedding provenance signals into NeRF representations, ensuring watermark persistence across novel views while enabling multiple uniquely keyed watermarks within a single model.  
\section{Preliminaries (TensoRF)}

This paper builds on TensoRF \cite{chen2022tensorf}, a popular explicit tensor-based representation for neural radiance fields (NeRFs). Unlike the traditional NeRF \cite{mildenhall2021nerf}, which relies on multi-layered perceptions (MLPs), TensoRF \cite{chen2022tensorf} represents a scene as a set of factorized tensors: the geometry grid, denoted by $G_{\sigma} \in \mathbb{R}^{I \times J \times K}$, which encodes the volume density $\sigma$ at each voxel in the 3D grid; and the appearance grid $G_{c} \in \mathbb{R}^{I \times J \times K \times P}$, which encodes the view-dependent color \textit{c}; where \textit{I}, \textit{J}, \textit{K} represent the resolutions of the feature grid along the X, Y, Z axes. \textit{P} denotes the number of appearance feature channels. Given a 3D location $\mathbf{x} = (x, y, z) \in \mathbb{R}^3$ and a view direction $\mathbf{d}$, by trilinear interpolation is applied to sample the two grids and the corresponding density $\sigma$ and color \textit{c} is estimated by: 
\begin{equation}
    \sigma, \mathbf{c} = \left( G_{\sigma}(\mathbf{x}), S(G_{\mathbf{c}}(\mathbf{x}), \mathbf{d}) \right)
\end{equation}

Where \textit{S} is a decoding function, the decoding function can be either a small MLP or Spherical Harmonic (SH) function that covers the appearance features and view direction to an RGB color. The trilinearly interpolated grids $G_{\sigma}$ and $G_{c}$ are represented as:


{\scriptsize
\begin{equation}
    \begin{split}
        G_{\sigma}(\mathbf{x}) &= \sum_{r} \sum_{m} \mathcal{A}^{m}_{\sigma, r}(\mathbf{x}), \\
        G_{\mathbf{c}}(\mathbf{x}) &= \mathbf{B}\Bigl( \textstyle\bigoplus \Bigl[ \mathcal{A}^{m}_{\mathbf{c}, r}(\mathbf{x}) \Bigr]_{m, r} \Bigr)
    \end{split}
\end{equation}
}



Where, $\mathcal{A}^{m}_{\sigma, r}(\mathbf{x})$ and $\mathcal{A}^{m}_{\mathbf{c}, r}(\mathbf{x})$ are factorized components of the density and appearance tensors, indexed by mode \textit{m} and rank \textit{r}. $\mathbf{B}$ matrix acts as a global appearance dictionary that captures correlations across the scene. While $\textstyle\bigoplus$ denotes the concatenation of tensor components.


\begin{figure*}[t]
    \centering
    \includegraphics[width=\textwidth, height=7cm]{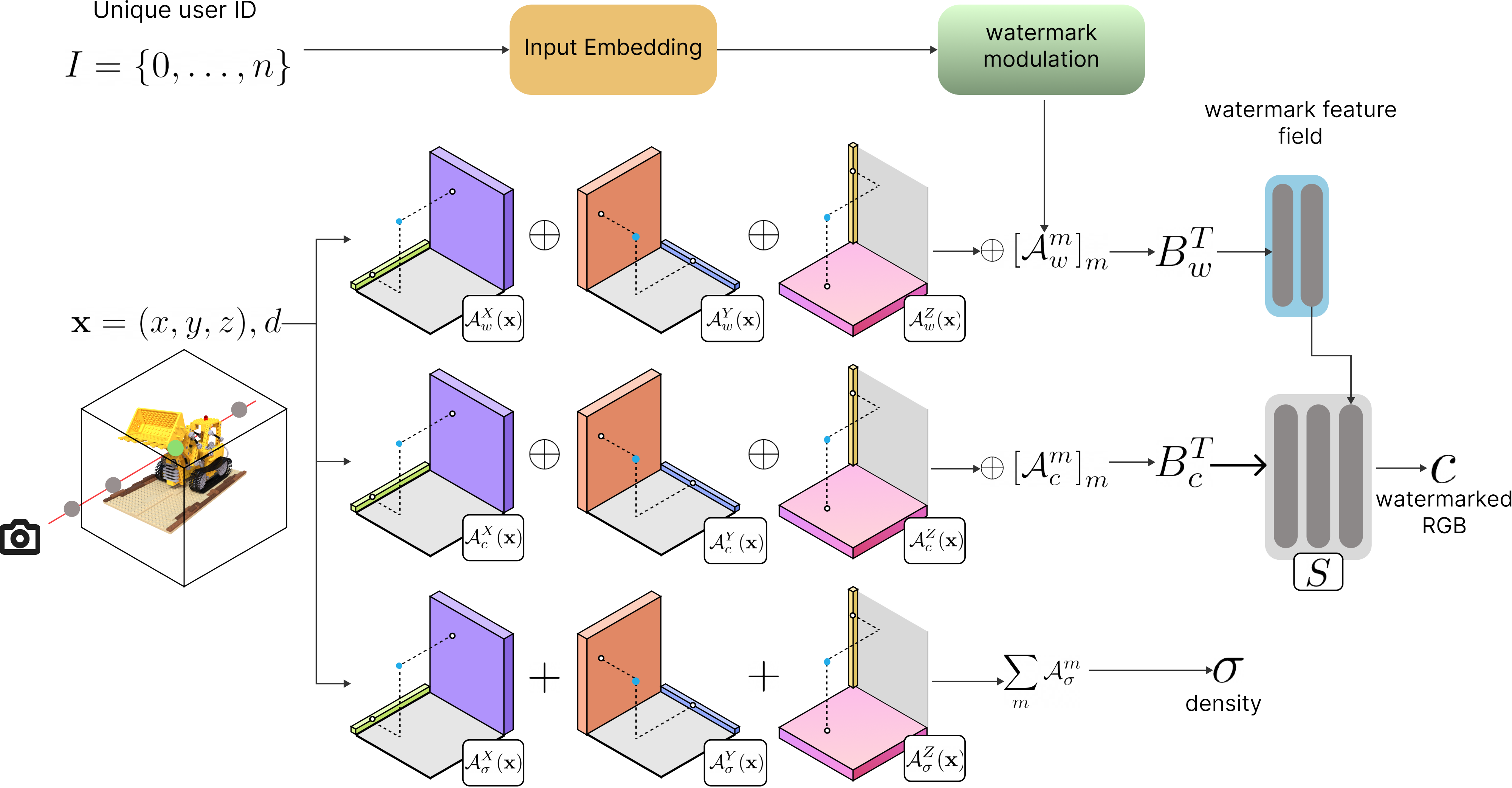}
    \caption{MultiNeRF extends TensoRF by introducing a watermark grid $A_w$ alongside the geometry $A_\sigma$ and appearance $A_c$ grids. Unique watermark IDs ($1..n$) are first encoded via a learnable embedding network into compact vectors that are then transformed into per-channel scaling ($\gamma$) and shifting ($\beta$) parameters. These parameters modulate the watermark grid’s features, ensuring that each distinct message is selectively activated and merged with the appearance grid during inference.}
    \label{fig:arch}
\end{figure*}

TensoRF employs differentiable volume rendering. It integrates the radiance values along the rays cast through the scene, which samples \textit{Q} points along the ray. Given a ray passing through a 3D point, the final pixel color \textit{C} is computed as:
\begin{equation}
    C = \sum_{q=1}^{Q} \tau_q \left( 1 - \exp(-\sigma_q \Delta_q) \right) c_q
\end{equation}
where, $\sigma_q$ and $c_q$ are the density and color of the sampled point $x_q$, $\Delta_q$ is the step size along the ray, and $\tau_q$ is the transmittance.
\begin{equation}
    \tau_q = \exp \left( - \sum_{p=1}^{q-1} \sigma_p \Delta_p \right)
\end{equation}
\section{Methodology}

\label{sec:method}

Since the TensoRF decomposes a scene into two separate grids for geometry and appearance, MultiNeRF extends this framework to encode multiple watermarks with a further dedicated grid.  In contrast to prior approaches that embed only a single watermark (\eg \cite{jang2024waterf, song2024protecting}), our method accommodates multiple distinct messages, each identified by a unique key (ID), which forms an additional model input. \cref{fig:arch} provides a high-level overview of the pipeline.

\subsection{Constructing the Watermarking module} \label{subsec:buildmodel}

As NeRF models only take position and view direction as inputs to the model, we need an input embedding layer to condition the NeRF and embed the watermark accordingly, allowing it to switch between messages. We let $I(n)$ be the integer IDs, each corresponding to a distinct watermark message $M(n)$, and introduce a learnable embedding layer $Emb()$ which maps $I(n)$ to $e_n$ a 16 dimensional message vector.
\begin{equation}
    e_n = \text{Emb}(I(n))
\end{equation}
$Emb$ is a learnable small MLP, and $e_n$ is the embedding vector, which will serve as a condition to modulate and select from multiple learned watermarks.

\textbf{Watermark Grid}. TensoRF \citep{chen2022tensorf} decomposes a scene into two explicit 3D grids: $G_{\sigma}$ for geometry (density) and $G_{\mathbf{c}}$ for appearance (color). Directly embedding the watermark into the existing appearance grid $G_{\mathbf{c}}$ would risk entangling the watermark features with the scene color, potentially degrading both.  We propose an additional grid $G_{\mathbf{w}}$ of the same spatial resolution ${(I \times J \times K)}$ but containing watermark-specific feature channels. Given, at any 3D location $\mathbf{x} = (x, y, z)$,
\begin{equation}
    w(\mathbf{x}) =\textstyle\bigoplus \left[ \mathcal{A}^{m}_{\mathbf{w}, r}(\mathbf{x}) \right]_{m, r}
\end{equation}

\textbf{Watermark modulation.} A design challenge in embedding multiple watermarks is ensuring that the model `activates' only the selected watermark's features. To this end, we draw inspiration from FiLM \citep{perez2018film}; its FiLM layer can influence the feature space computation using feature-wise affine transform based on external information. In our context, as we want to embed multiple watermarks, our external information is the embedded watermark ID, i.e., $e_n$. Our watermark modulation has a modulator network, which in our case is a learnable linear layer $Mod()$, which takes $e_n$ as input and outputs a pair of modulation parameters to scale and shift the feature space of the watermark features: scale $\gamma_n$ and shift $\beta_n$:
\begin{equation}
    [\gamma_n, \beta_n] = \text{Mod}(\mathbf{e}_n),
\end{equation}
Thus, each $w(x)$ channel is conditioned based on the embedding  $e_n$. We then compute:
\begin{equation}
    \mathbf{w'}(\mathbf{x}) = \gamma_n \odot \mathbf{w}(\mathbf{x}) + \beta_n,
\end{equation}
Where $\odot$ is the Hadamard product. The resulting $\mathbf{w'}(\mathbf{x})$ represents the \textit{modulated watermark features} specific to $I(n)$. Intuitively, $\gamma_n$ and $\beta_n$ activate certain dimensions of the watermark grid differently for each watermark ID.

\textbf{Merging watermark and appearance.} To incorporate these modulated watermark features with appearance, we pass only $G_{c}(\mathbf{x})$ and viewing direction to the S MLP, which is a color decoding function of TensoRF (see \cite{chen2022tensorf} for details), and we apply the modulated watermark features to the last linear layer of the S MLP:
\begin{equation}
    \tilde{\mathbf{c}}(\mathbf{x}) = S(\mathbf{G}_c(\mathbf{x}), \mathbf{w'}(\mathbf{x}), \mathbf{d}),
\end{equation}
where $\tilde{\mathbf{c}}(\mathbf{x})$ is now a watermark color at location $\mathbf{x}$. The rest of TensoRF's volume rendering proceeds unchanged.
Because each ID yields different $\{\gamma_n, \beta_n\},$, the final color $\tilde{\mathbf{c}}(\mathbf{x})$ implicitly contains an ID-specific watermark pattern. As a result, any novel view rendered from the watermarked NeRF model will contain an embedded signature that can be extracted with the watermark decoder.

\textbf{Differentiable Augmentation layer}. To promote robustness against image attacks, we train with differentiable augmentations on the fully rendered image. Each image is transformed by a small set of augmentations (brightness, contrast, color jiggle, gaussian blur, gaussian noise, hue, posterize, RGB shift, saturation, median blur, box blur, motion blur,  sharpness, and differentiable JPEG compression) before passing them into the watermark decoder function.  Formally, if $\tilde{\mathbf{I}}$ is the full rendered image, we apply a random pair of augmentations from $\mathbf{A}$ from a set of $\{A_1,...,A_m\}$.:
\begin{equation}
    \hat{\mathbf{I}} = \mathcal{A}(\tilde{\mathbf{I}}),
\end{equation}
where $\hat{\mathbf{I}}$ is an augmented image, and we pass this image to the watermark decoder.

\subsection{Training MultiNeRF to embed watermarks} 
\label{subsec:finetune}

\begin{figure}[h]  
    \centering
    \includegraphics[width=\linewidth]{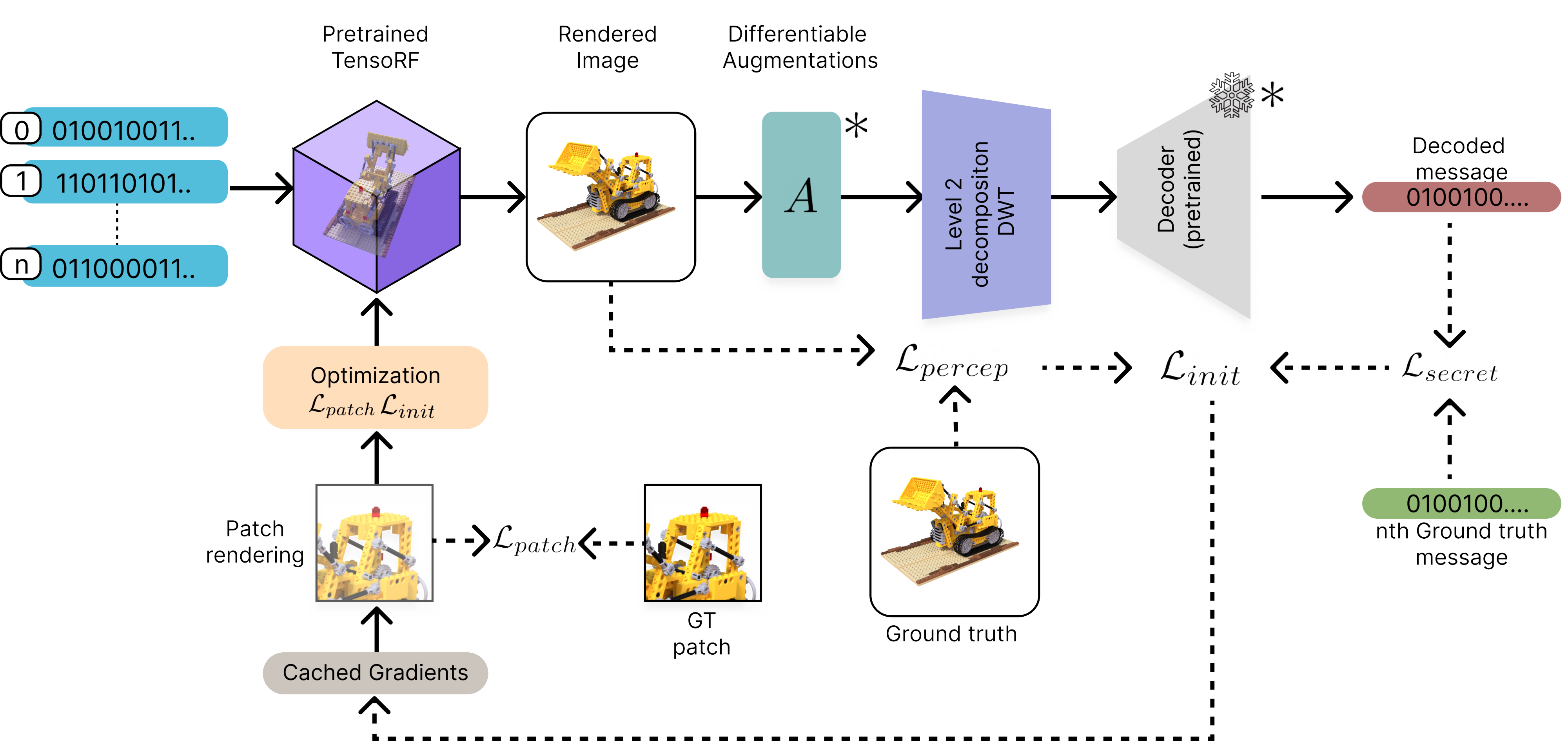} 
    \caption{MultiNeRF training process. A learnable encoder transforms multiple distinct watermark IDs into compact embedding vectors. MultiNeRF is trained with an end-to-end HiDDEN decoder module that renders images passing through differentiable noise augmentations. Perceptual (\cref{eq:init}) and patch-based (\cref{eq:patch}) reconstruction losses balance quality against accuracy (\cref{eq:bce}).}
    \label{fig:arch2}
\end{figure}


We begin by training a TensoRF NeRF model, using the geometry and appearance grids to initialize those parts of our MultiNeRF architecture.  We perform a separate initialization of the watermark decoder training a HiDDeN decoder following \citep{zhu2018hidden}. Other parts of our architecture are initialized with white noise, and training proceeds in two phases.

\textbf{Phase 1.} We begin by first freezing the watermark decoder $D$. We generate a random watermark ID $I(n)$, with $\mathbf{m}_n$ the ground truth message for $n-th$ ID. For each training iteration, we render a full-resolution image $\mathbf{I} \in \mathbb{R}^{H \times W \times 3}$. As directly injecting watermarks into the spatial domain can be vulnerable to image compression, we apply a level 2 decomposition of Discrete Wavelet Transform (DWT) on the rendered image (after \cite{jang2024waterf}) and use the $LL_2$ sub-band as it retains the majority of the energy and broad structural features. This sub-band image (denoted $\mathbf{I}_{LL_2}$) is then passed to the watermark decoder $D$ obtaining a decoded message $\mathbf{m}'_n$.  A BCE loss is calculated between secrets $\mathbf{m}_n$ and $\mathbf{m}'_n$:
\begin{equation}
    \mathcal{L}_{secret} = \text{BCE}(\mathbf{m}_n,\mathbf{m}'_n) \label{eq:bce}
\end{equation}
Alongside the secret loss $\mathcal{L}_{secret}$, we also calculate a perceptual loss $\mathcal{L}_{percept}$ via Watson-VGG \citep{czolbe2020loss} to create a total loss for this first phase of training:
\begin{equation}
    \mathcal{L}_{init} = \lambda_m \mathcal{L}_{secret} + \lambda_i \mathcal{L}_{percept}  \label{eq:init}
\end{equation}

\textbf{Phase 2.} Training continues end-to-end via a patch-based process using deferred backpropagation \citep{zhang2022arf}.  As optimizing at a full resolution in each iteration can be memory-intensive. Therefore, we make patches and re-render the images from those patches. A patch loss $\mathcal{L}_{patch}$ is then calculated, which uses the RGB loss across the rendered pixels, SSIM loss, and total variation regularization:
\begin{equation}
    \mathcal{L}_{patch} = \lambda_{rgb} \mathcal{L}_{RGB} + \lambda_{TV} \mathcal{L}_{TV} + \lambda_{SSIM} \mathcal{L}_{SSIM} \label{eq:patch} 
\end{equation}
We unfreeze the decoder $D$ and introduce the differentiable augmentations $A$ at each iteration on the rendered image, on which the 2-level DWT is applied and passed into the watermark decoder $D$.  

\section{Experiments and Discussion}
\label{sec:eval}

\subsection{Experimental Setup} 

\textit{\textbf{Datasets:}} We train and evaluate our method using the Nerf-Synthetic \cite{mildenhall2021nerf} (hereafter, SYN) and the LLFF datasets \cite{mildenhall2019local}. SYN consists of eight representative scenes: {\em chair, drums, ficus, hotdog, lego, materials, mic, ship}; and the LLFF dataset consists of eight scenes: {\em fern, flower, fortress, horns, leaves, orchids, room, trex}.  In all evaluations, we train and test MultiNeRF using the partitions of those public datasets.

\textit{\textbf{Baselines:}} As no multiple watermark frameworks are available for NeRFs, we evaluate our proposed method against two state-of-the-art NeRF watermarking methods for the single watermarking task: WateRF \citep{jang2024waterf} and NeRFProtector \citep{song2024protecting}. For fair comparisons, we use the models that those baselines have implemented in their methods; for WateRF, we choose TensoRF \citep{chen2022tensorf} as the NeRF model, and for NeRFProtector \citep{song2024protecting} we use Instant-NGP \citep{muller2022instant}. As no prior work embeds multiple watermarks into a single NeRF, we tune WateRF \citep{jang2024waterf} by adding an input embedding layer to condition it to embed multiple watermarks (denoted `WateRF-modified').

\textit{\textbf{Metrics:}} We use bit accuracy to evaluate the accuracy of the decoder for a given capacity.  For visual quality, we measure  PSNR, SSIM \citep{wang2004image}, and LPIPS \citep{zhang2018unreasonable} distances between the ground truth and watermarked image.

\textit{\textbf{Training setup:}} MultiNeRF is implemented in PyTorch and trained with a batch size of 1 on a single NVIDIA RTX 4080. The ADAM optimizer \citep{kingma2014adam} is used with $\beta_1 = 0.9$ and $\beta_2 = 0.99$, and an exponential learning rate decay is applied. All the grid parameters are optimized with an initial learning rate of 0.02, and the basis matrix for all grids has a learning rate of $1e-3$. The $\lambda_i$ and $\lambda_m$ in \cref{eq:init} is set to 0.05 and 0.95; while for the patch loss, in \cref{eq:patch}, we set $\lambda_{RGB}$, $\lambda_{TV}$ and $\lambda_{SSIM}$ to 0.1, 0.02, and 0.20 respectively. 

\subsection{Single watermark embedding evaluation}
We evaluate two variants of our model trained with (MultiNeRF-Noised) and without (MultiNeRF) noise augmentations. All methods embed a single 48-bit message.  As the bit accuracy is slightly affected by different messages, we average over 50 sets of unique watermarks and report the average results for all the synthetic and LLFF datasets. We enforce a minimum Hamming distance between the watermark messages to ensure embedded messages are distinct.  Note that NeRFProtector reported on only 3 scenes from SYN and LLFF datasets \citep{song2024protecting}.  Using the official implementation, we cannot reconstruct 3 of the scenes {\em (ficus, flower, leaves)} and omit these outliers from their average.

\cref{tab:bit_acc_single} shows that MultiNeRF achieves a mean bit accuracy of 93.18\% on SYN outperforming both WateRF (91.51\%) and NeRFProtector (90.81\%); and like WaterRF saturates performance on LLLF at $\sim 99\%$.  Whilst the noised variant (MultiNeRF-Noised) exhibits slightly lower bit accuracies in some scenes, this is traded for improved robustness (c.f. \cref{subsec:robust}). MultiNeRF generally achieves comparable quality and slightly higher accuracy at the single watermarking task.

\begin{table}[h]
    \centering
    \setlength{\tabcolsep}{2pt}
    \resizebox{\columnwidth}{!}{%
    \begin{tabular}{l|ccccccccc}
        \toprule
        Method  (on SYN) & Avg. & Chair & Drums & Ficus & Hotdog & Lego & Materials & Mic & Ship \\
        \midrule
        WateRF \citep{jang2024waterf} & 91.51 & 98.31 & 92.19 & 79.83 & 96.21 & 93.16 & 82.33 & 95.92 & 94.10 \\
        NeRFProtector \citep{song2024protecting} & 90.81 & 96.41 & 89.73 & - & 93.47 & 90.12 & 84.05 & 90.39 & 91.54 \\
        \textbf{MultiNeRF (ours) }& 93.18 & 98.35 & 95.14 & 83.06 & 96.97 & 94.86 & 85.16 & 96.89 & 95.03 \\
        \textbf{MultiNeRF-Noised (ours)} & 89.70 & 92.60 & 93.61 & 78.60 & 94.36 & 92.49 & 83.54 & 89.72 & 92.65 \\
        \midrule
        Method (on LLFF)& Avg. & Fern & Flower & Fortress & Horns & Leaves & Orchids & Room & Trex \\
        \midrule
        WateRF \citep{jang2024waterf} & 99.32 & 99.75 & 99.56 & 99.95 & 99.92 & 99.53 & 96.07 & 99.89 & 99.91 \\
        NeRFProtector \citep{song2024protecting} & 95.73 & 94.68 & - & 99.58 & 98.77 & - & 82.23 & 99.73 & 99.37 \\
        \textbf{MultiNeRF (ours)}& 99.23 & 99.39 & 99.48 & 99.82 & 99.87 & 99.68 & 95.92 & 99.77 & 99.88 \\
        \textbf{MultiNeRF-Noised (ours)} & 98.55 & 99.04 & 99.05 & 99.90 & 99.86 & 99.28 & 91.81 & 99.65 & 99.81 \\
        \bottomrule
    \end{tabular}
    }
    \caption{Comparing raw bit accuracy $\uparrow$ (no error correction) of the proposed method (MultiNeRF) to baseline methods (NeRFProtector, WaterRF) for the single message task on datasets: SYN (upper) and LLFF datasets (lower). Values to 2 d.p.}
    \label{tab:bit_acc_single}
\end{table}

\begin{figure*}[t]
    \centering
    \includegraphics[width=\textwidth, height=12cm]{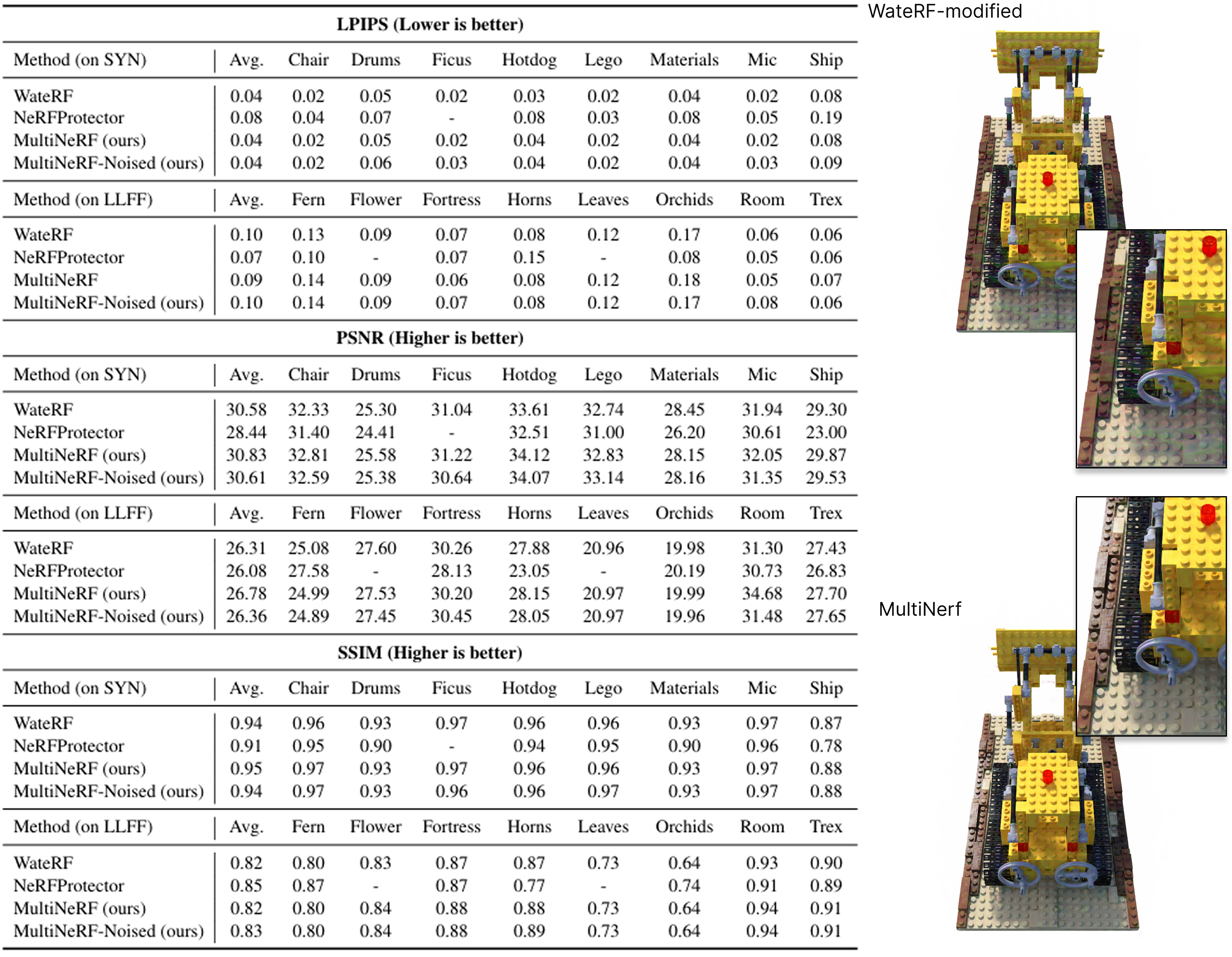}
    \caption{Comparing visual quality.  Left: Table of metrics (LPIPS $\downarrow$, PSNR $\uparrow$, SSIM $\uparrow$) of MultiNeRF to baselines (NeRFProtector, WateRF) for single message task on SYN and LLFF datasets. Values to 2 d.p. b) Right: Examples of visual artifacts (colored ripples) present in WateRF-modified versus the proposed MultiNeRF method (zoom inset).}
    \label{fig:quality_and_visual_artifacts}
\end{figure*}

\cref{fig:quality_and_visual_artifacts} shows that quality is equivocal for the baselines versus MultiNeRF, which attains an average PSNR of 30.83 dB for SYN and 26.78 dB for the LLFF dataset. We are slightly higher (SSIM, PSNR) or lower (LPIPS) than our baselines on some metrics. Overall, we conclude that MultiNeRF is comparable in quality and slightly outperforms bit accuracy for the single watermarking task.

\subsection{Evaluating multiple watermark embedding}
\label{subsec:multiple_wm}
We evaluate the model setting of embedding multiple distinct watermarks within a single NeRF model. Since all baselines target a single watermark per model, we have adapted WateRF to accept an input variable to select a watermark and trained it as a baseline (WateRF-modified). Each watermark has a 16-bit capacity, and we embed multiple watermarks into a single model.

\cref{fig:bit_acc_mwm} presents the average bit accuracy across the SYN and LLFF datasets for varying numbers of unique watermarks per model.  For SYN, WateRF-modified returns a random response (50\% bit accuracy) for 32 watermarks and onwards versus MultiNeRF at 70\%  bit accuracy on 32 watermarks, which degrades to a random response at 64 watermarks. Similarly, for LLFF, we observe that both methods start with similar bit accuracies at  2 watermarks, with WateRF-modified dropping to random response at 32 watermarks, whilst MultiNeRF achieves $82\%$ bit accuracy at 32 watermarks and $68\%$  at 64 watermarks.  The error bars confirm a significant bit accuracy improvement using MultiNeRF for the multiple watermarking task (\ie $p<0.0001$ for both SYN and LLFF).

\begin{figure}[t!]
\begin{center}
\includegraphics[width=1.0\linewidth,height=7cm]{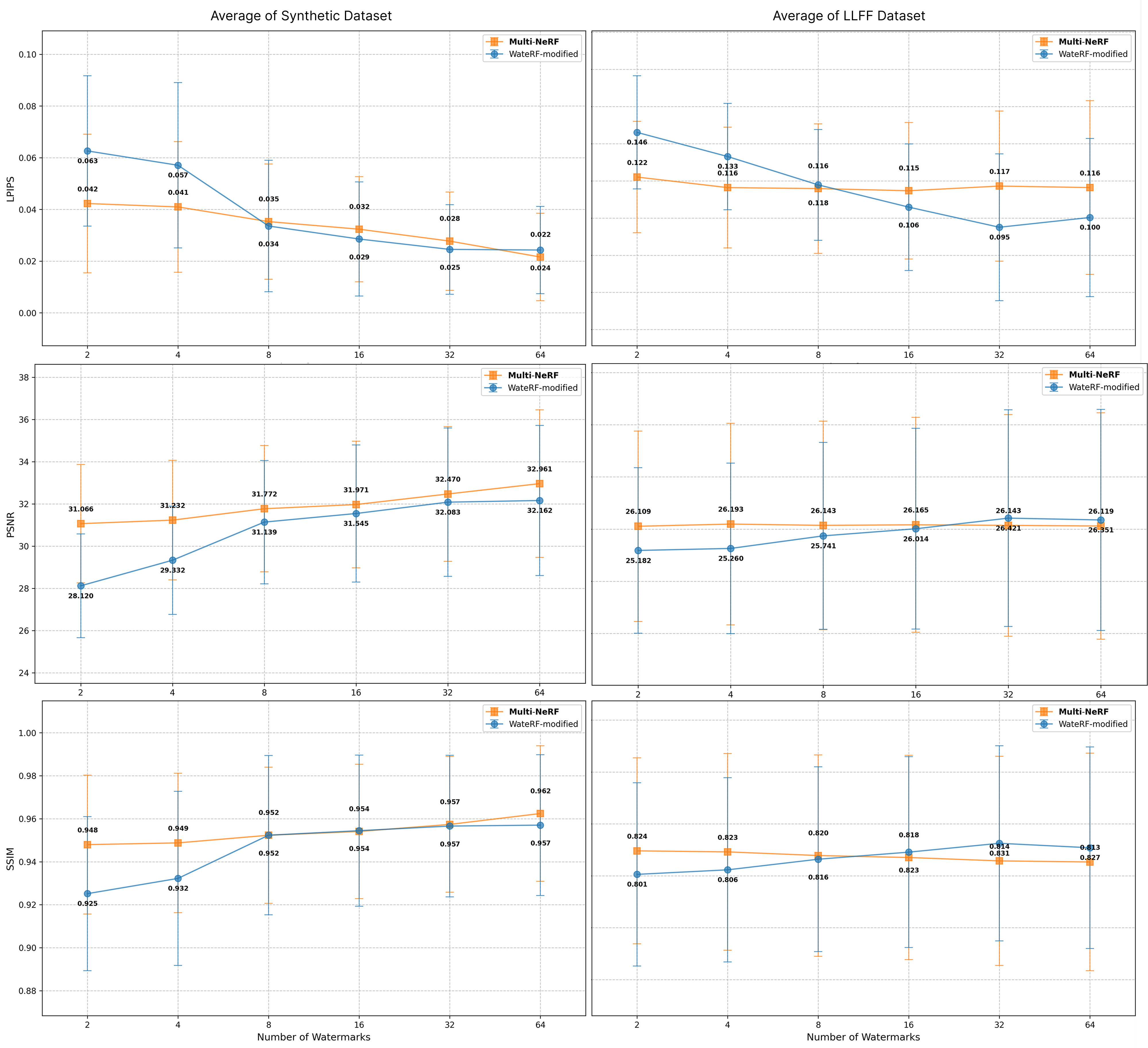}
\end{center}
\caption{Evaluating the visual quality of MultiNeRF vs. baseline WateRF-modified for the multi-watermarking task: LPIPS (top); PSNR (mid.); SSIM (bot.).}
\label{fig:quality_mwm} 
\end{figure}

\begin{figure*}[t!]
\begin{center}
\includegraphics[width=1\linewidth]{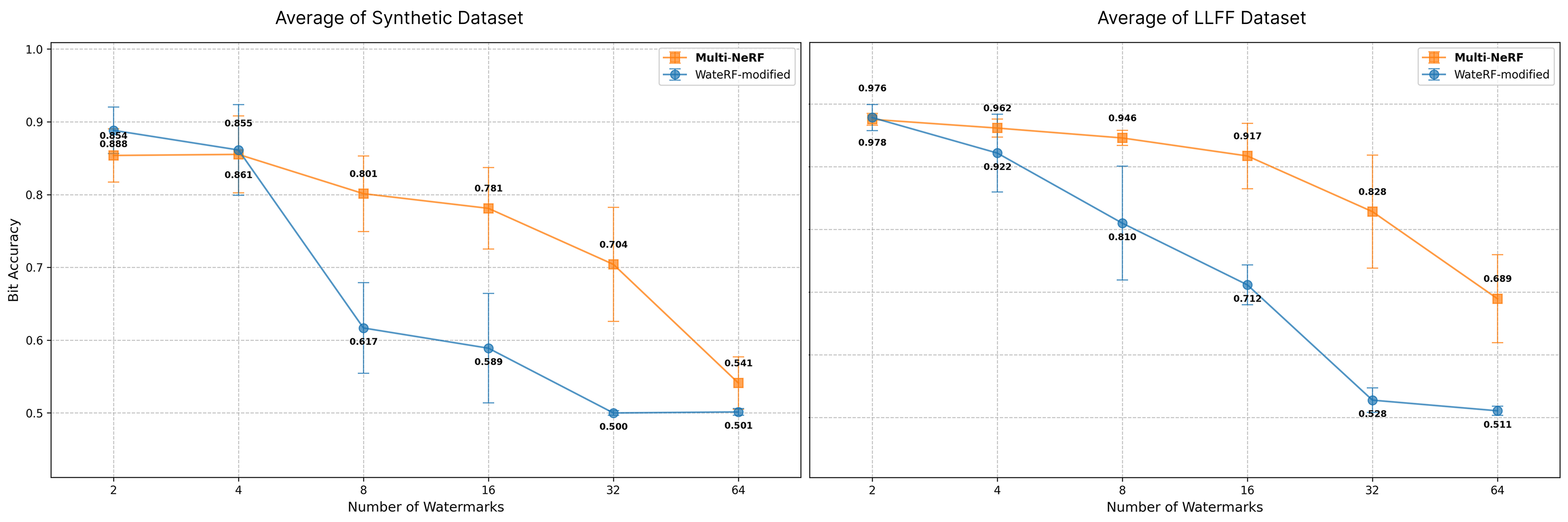}
\end{center}
\caption{Evaluating bit accuracy $\uparrow$ of MultiNeRF versus baseline WateRF-modified for the multi-watermarking task. Accuracies averaged for SYN (left) and LLFF (right) datasets. Performance is significantly higher for MultiNeRF beyond the single watermarking case.}
\label{fig:bit_acc_mwm} 
\end{figure*}

\cref{fig:quality_mwm} shows all methods performed with similar visual quality, and the error bars indicate no statistically significant difference in quality between the methods. \cref{tab:ablation_selected} illustrates the trade-off between model size and performance for 16 watermarks. Compared to TensoRF \citep{chen2022tensorf}, introducing a watermark grid increases parameter count and storage overhead by $\sim 12\%$.


\begin{figure}[t!]
\begin{center}
\includegraphics[width=1.0\linewidth, height=4cm]{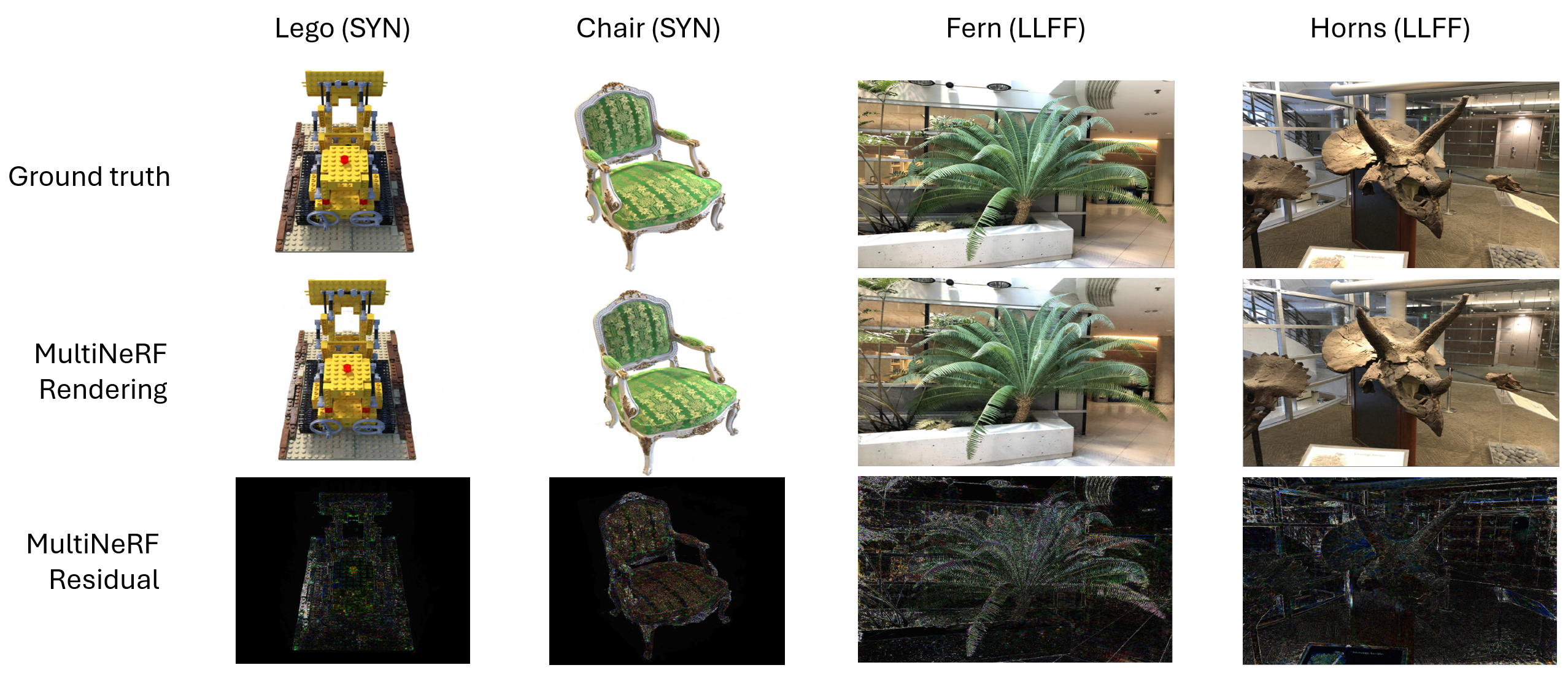}
\end{center}
\label{fig:residual} 
\caption{Visualizing the imperceptible watermark residual (bottom, amplified) introduced by MultiNeRF between the ground truth (top) and watermarked image (middle). }
\end{figure}

\subsection{Evaluating watermark robustness}
\label{subsec:robust}

\begin{figure}[t!]
\begin{center}
\includegraphics[width=1\linewidth]{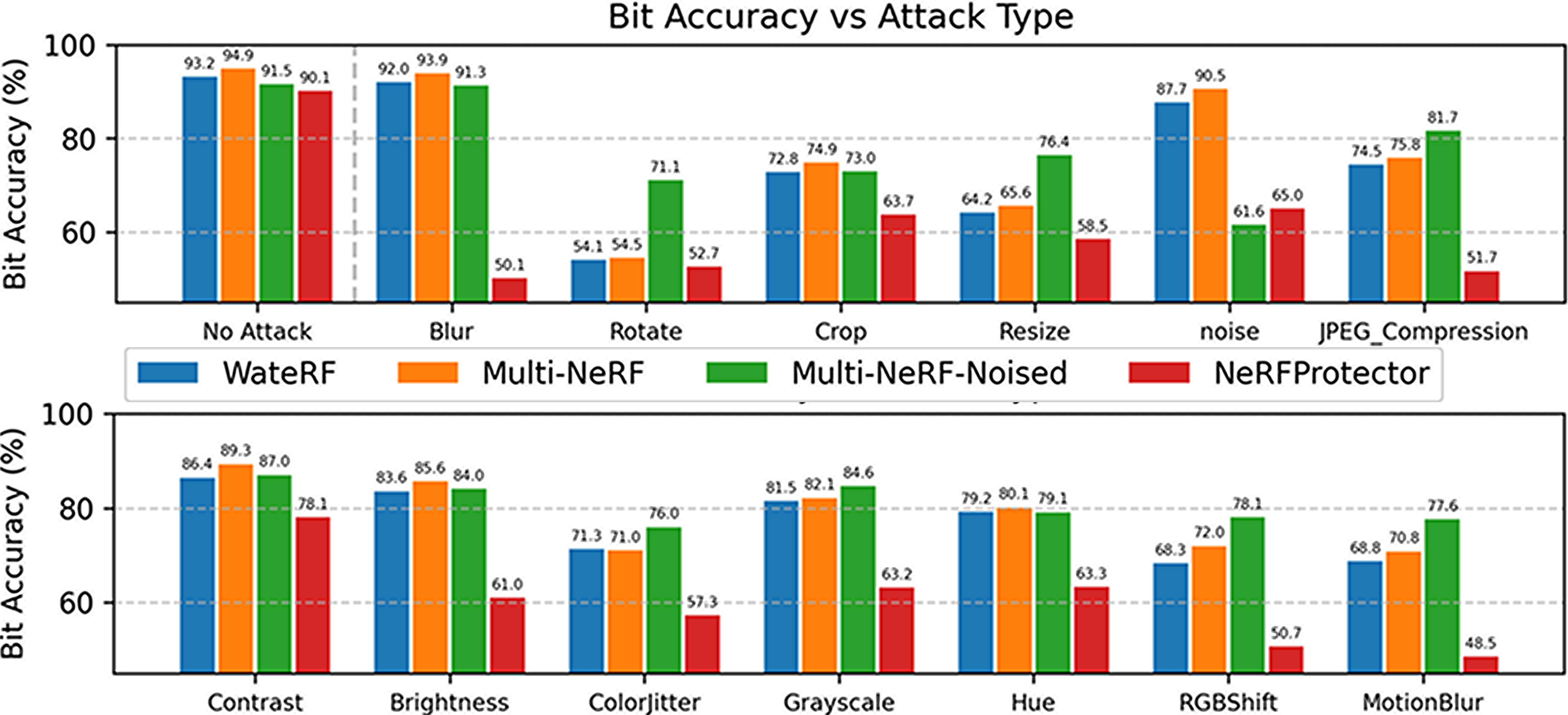}
\end{center}
\caption{Evaluating the robustness to various degrading transformations for the proposed MultiNeRF method trained with (-noised) and without noise augmentations to baselines WateRF and NeRFProtector for watermarking a single message (averaged for 50 runs over Lego).}
\label{fig:robustness} 
\end{figure}

\begin{table}[t!]
    \centering
    \scriptsize
    \renewcommand{\arraystretch}{1.1}
    \setlength{\tabcolsep}{2pt} 
    \resizebox{\columnwidth}{!}{%
    \begin{tabular}{c|cccc}
        \toprule
         & No Attack & Diffusion  \cite{zhao2024invisible} & VAE-Cheng \cite{cheng2020learned} & VAE-Bmshj \cite{balle2018variational} \\
        \midrule
        WateRF & 89.98 & 64.68 & 56.78 & 56.96 \\
        NeRFProtector & 78.59 & 51.56 & 63.35 & 63.00 \\
        \textbf{MN} & 91.51 & 67.53 & 56.26 & 56.48 \\
        \textbf{MN-noised} & \textbf{92.52} & \textbf{80.48} & \textbf{73.94} & \textbf{74.72} \\
        \bottomrule
    \end{tabular}}
    \caption{Comparison of bit accuracies (persistence) under various regeneration attacks.}
    \label{tab:regeneration_attack}
\end{table}
Robustness evaluation was performed on the SYN/Lego scene, and results averaged for 50 different watermarks embedded into the model. In this experiment, we used standard image transformations such as cropping, blur, and JPEG compression and the regeneration attacks proposed in \citep{zhao2024invisible}. 
\cref{fig:robustness} plots the bit accuracy vs the types of attacks used. The MultiNeRF-Noised model has greater bit accuracy on these attacks vs. baselines and a 3\% bit lower accuracy than MultiNeRF.

\cref{tab:regeneration_attack} compares the bit accuracy of MultiNeRF and the baselines under three specific regeneration attacks. We test one diffusion based attack (\textbf{Diffusion} \cite{zhao2024invisible} while the other two are VAE based regeneration attacks (\textbf{VAE-Cheng} \cite{cheng2020learned}, and \textbf{VAE-Bmshj} \cite{balle2018variational}). Notably, the MN-noised variant exhibits superior robustness, maintaining a high bit accuracy under regeneration attacks. These results indicate that while other methods experience bit accuracy degradation under the image transformation and regeneration attacks, introducing the Differentiable Augmentation Layer (in MN-noised) increases the overall robustness.

\subsection{Ablation Study}

\textbf{Watermark Grid.} Here, we present ablation experiments to validate the design choices in MultiNeRF (\cref{tab:ablation}) conducted on the Lego scene with 16 watermarks per model.  First, we remove the proposed watermark grid (-GRID), which falls back to an MLP to encode the watermark information. Without a separate spatial grid, bit accuracy drops to 51.60\%, suggesting that an explicit grid provides higher capacity for multiple watermarks. We also examine the effect of removing modulation (-FiLM), causing a drop from 82.79 to 70 \%.  We explored a training simplification where the watermark grid is learned without adjusting the appearance or geometry grid ($G_{\sigma}$ frozen), showing that bit accuracy suffers when we do not allow the appearance parameters to adapt. Another variation injects the FiLM-modulated features into earlier layers (+EARLY) of MLP rather than the last layer. While the bit accuracy (78.04\%) remains high, visual quality suffers \eg as with WateRF-modified, scene reflections are absent.

\begin{table}[t!]
    
    \centering
    \scriptsize
    \renewcommand{\arraystretch}{1.1}
    \setlength{\tabcolsep}{4pt}
    
    \begin{subtable}[t]{0.48\textwidth}
        \centering
        \caption{Ablation Study}
        \begin{tabular}{l|cccc}
            \toprule
            Method & Bit Acc. $\uparrow$ & PSNR $\uparrow$ & LPIPS $\downarrow$ & SSIM $\uparrow$  \\
            \midrule
            MN(-GRID)(+MLP) & 51.60 & 35.50 & 0.01 &  0.98  \\
            MN(-FiLM)(-GRID) & 66.28 & 33.57 & 0.02 &  0.97  \\
            MN(-FiLM) & 70.00 & 34.69 & 0.01 & 0.97  \\
            MN $G_{\sigma}$ frozen & 67.72 & 34.56 & 0.01  & 0.97 \\
            MN(+EARLY) & 78.04 & 34.29 & 0.02  & 0.97  \\
            MN(-GRID) & 66.35 & 33.79 & 0.02  & 0.97 \\
            \textbf{MN (ours)} & \textbf{82.79} & 34.03 & 0.01 & 0.97  \\
            \bottomrule
        \end{tabular}
        \label{tab:ablation}
    \end{subtable}
    \hfill
    \begin{subtable}[t]{0.48\textwidth}
        \centering
        \caption{Performance vs. Size}
        \begin{tabular}{l|ccc}
            \toprule
            Method & Bit Acc. $\uparrow$ & Size (MB) $\downarrow$ & Params (M) \\
            \midrule
            TensoRF  & - & 68.6 & 17.75 \\
            MN(-GRID)(-FiLM) & 66.28 & 68.8 & 17.76 \\
            MN (ours) & \textbf{82.79} & 77.1 & 19.98 \\
            \bottomrule
        \end{tabular}
        \label{tab:ablation_selected}
    \end{subtable}
    \caption{Top: Ablation Study of MultiNeRF (abbrev. MN), evaluating the impact of key components. Bottom: Comparison of performance gain (accuracy) versus size increase. Figures to 2.d.p.}
\end{table}

\begin{table}[t!]
    
    \centering
    \scriptsize
    \renewcommand{\arraystretch}{1.1}
    \setlength{\tabcolsep}{4pt}
    
    \begin{tabular}{c|ccccc}
        \toprule
        \#WMCom & Bit Acc. $\uparrow$ & PSNR $\uparrow$ & LPIPS $\downarrow$ & SSIM $\uparrow$ & Param-WMGrid(M) \\
        \midrule
        2  & 78.91 & 34.00 & 0.01 & 0.97 & 0.5 \\
        4  & 79.77 & 34.08 & 0.01 & 0.97 & 1.1 \\
        8  & 82.79 & 34.03 & 0.01 & 0.97 & 2.2 \\
        16 & 82.83 & 34.00 & 0.01 & 0.97 & 4.4 \\
        24 & 85.66 & 33.92 & 0.02 & 0.97 & 6.6 \\
        32 & 85.40 & 33.89 & 0.02 & 0.97 & 8.8 \\
        48 & 86.03 & 33.85 & 0.02 & 0.97 & 13.2 \\
        \bottomrule
    \end{tabular}
    \caption{Ablation Study on the Number of Watermark Components, Evaluating the Impact on Performance Metrics.}
    \label{tab:tensor_components_ablation}
\end{table}

\textbf{Number of Tensor Components.} The number of tensor components for watermark grid (\#WMCom) refers to the rank of the factorized tensor as described in \cite{chen2022tensorf}. We fix 16 and 48 components for the density and appearance grids while varying \#WMCom. From \cref{tab:tensor_components_ablation}, we observe an upward trend in accuracy as we increase \#WMCom, reaching a peak at 48 components. However, this comes at the cost of decreasing quality (PSNR and LPIPS). Both 8 and 16 have comparable bit accuracy $\sim 82.8\%$, and as 16 components offers only marginal accuracy gain over 8 whilst doubling model parameter count, we selected 8 components as the optimal trade-off between robustness and quality.

\textbf{Scalability across different bit-length decoder.} We compare different bit-length decoders trained on 16 watermarks per scene (\cref{tab:ablation_bit_length_sclable}).  Both quality and accuracy somewhat invariant to different bit-length decoders for this task, showing the flexibility of our method, which adapts seamlessly to varying bit payload size. 

\textbf{User Study on Watermark Artifacts.} To evaluate the perceptual quality of our method, we conducted a user study where participants rated watermark artifacts on a scale from 1 (low artifacts) to 5 (high artifacts). \cref{tab:artifacts} shows the mean artifact scores, standard deviations and statistical significance (p-values) from t-tests comparing each method to MultiNeRF. Our method achieves one of the lowest artifact scores (3.34), comparable to WateRF \cite{jang2024waterf} (3.46, $p > 0.05$), indicating no statistically significant difference. However, compared to WateRF-modified (3.65, $p < 0.05$) and NeRFProtector \cite{song2024protecting} (3.61, $p < 0.05$), our method shows a statistically significant reduction in artifacts, confirming that it introduces fewer perceptible watermark distortions than the other baselines. This demonstrates that MultiNeRF effectively balances watermark robustness with minimal perceptual degradation.

\begin{table}[t!]
    
    \centering
    \scriptsize
    \renewcommand{\arraystretch}{1.1}
    \setlength{\tabcolsep}{4pt}

    \begin{tabular}{c|cccc|cccc}
        \toprule
        & \multicolumn{8}{c}{\textbf{Decoder Bit Length}} \\
         \cmidrule(lr){2-9}
         & \multicolumn{4}{c|}{\textbf{Synthetic}} & \multicolumn{4}{c}{\textbf{LLFF}} \\
         & \textbf{8} & \textbf{16} & \textbf{32} & \textbf{48} & \textbf{8} & \textbf{16} & \textbf{32} & \textbf{48} \\
        \midrule
        \textbf{Bit Acc.} & 81.7 & 82.00 & 83.50 & 84.50 & 94.14 & 91.10 & 91.71 & 91.87 \\
        \textbf{PSNR} & 34.51 & 34.73 & 33.79 & 33.96 & 27.11 & 27.63 & 28.54 & 27.57 \\
        \textbf{LPIPS} & 0.02 & 0.02 & 0.03 & 0.03 & 0.14 & 0.11 & 0.11 & 0.11 \\
        \textbf{SSIM} & 0.97 & 0.97 & 0.96 & 0.95 & 0.83 & 0.85 & 0.84 & 0.84 \\
        \bottomrule
    \end{tabular}
    \caption{Comparison of different decoder bit length metrics for SYN and LLFF datasets.}
    \label{tab:ablation_bit_length_sclable}
\end{table}

\begin{table}[t!]
    
    \centering
    \scriptsize
    \renewcommand{\arraystretch}{1.1}
    \setlength{\tabcolsep}{4pt}
    
    \begin{tabular}{c|cccc}
        \toprule
        Method & Artifacts (Mean) $\downarrow$ & SD & p-value \\
        \midrule
        WateRF-modified & 3.65 & 1.16 & $p < 0.05$  \\
        NerfProc & 3.61 & 1.29 & $p < 0.05$  \\
        WateRF & 3.46 & 1.28 & $p > 0.05$  \\
        \textbf{MultiNeRF} & \textbf{3.34} & \textbf{1.35} & -  \\
        \bottomrule
    \end{tabular}
    \caption{User Study results: mean Artifact Score, Standard Deviation, and p-value for Statistical Significance.}
    \label{tab:artifacts}
\end{table}

\section{Conclusion}

We presented \textbf{MultiNeRF}, the first NeRF watermarking technique to embed multiple conditional watermarks simultaneously within a NeRF (TensoRF) model.  MultiNeRF achieves competitive watermark embedding performance at single message watermarking while introducing the novel capability of multiple watermark support. This increases model capacity because the watermark grid enables multiple messages to be simultaneously encoded without degrading the visual quality. FiLM-based modulation enables selection between the distinct watermarks at rendering time.  Future work could explore learnable perceptual metrics or NeRF-specific benchmarks to help better quantify artifacts introduced by watermark embedding. As NeRFs continue to evolve and find new applications, frameworks like MultiNeRF will play an essential role in securing the intellectual property rights of 3D content creators. Their integration with emerging media provenance standards presents another future direction.
{
    \small
    \bibliographystyle{ieeenat_fullname}

}

\end{document}